\documentclass[10pt, journal, twocolumn]{IEEEtran}

\usepackage{algorithm}
\usepackage{algorithmic}
\usepackage{bm}
\usepackage{colortbl}
\usepackage{arydshln}
\usepackage{multirow}
\usepackage{multicol}
\usepackage{xcolor}
\usepackage{caption}
\usepackage{url}            % simple URL typesetting
\usepackage{adjustbox}

\usepackage{epsfig}
\usepackage{amsmath,amsthm,amssymb,amsfonts}
\usepackage{bm}

\usepackage{color}

\usepackage{fancyhdr}

\usepackage{lipsum}  % This package provides dummy text, you can remove it in your actual document.

\fancypagestyle{firstpage}{
  \fancyhf{}  % Clear all header and footer fields
  \fancyfoot[C]{4th ACM International Conference on AI in Finance@ICAIF2023}  % Add your conference info
}

\begin{document}\sloppy

% Example definitions.
% --------------------
\def\x{{\mathbf x}}
\def\L{{\cal L}}

% Title.
% ------
\title{Enhancing Financial Sentiment Analysis via \\Retrieval Augmented Large Language Models}
%
% Single address.
% ---------------
\author{Boyu Zhang$^{1}$\textsuperscript{*},
        Hongyang (Bruce) Yang$^{2}$\textsuperscript{*}, Tianyu Zhou$^{3}$\textsuperscript{*}, 
        \\ Ali Babar$^{1}$,
        Xiao-Yang Liu$^{2}$\textsuperscript{\textsection},
        \\
$^1$The University of Adelaide\\
$^2$Columbia University\\
$^3$Brown University\\
Email: boyu.zhang01@adelaide.edu.au,  
         hy2500@columbia.edu,   
         \\zhoutianyu0426@gmail.com,
         ali.babar@adelaide.edu.au,
         xl2427@columbia.edu        
}
\maketitle
\begingroup\renewcommand\thefootnote{*}
\footnotetext{Equal contribution.}

\begingroup\renewcommand\thefootnote{\textsection}
\footnotetext{Corresponding author.}
\endgroup

\maketitle

% Apply the custom page style for the first page
\thispagestyle{firstpage}

% Your paper starts here
%\lipsum[1-3] % Replace this with your actual text

% ... rest of your paper ...

%%%%%%%%%%%%%%%%%%%%%%%%%%%%%%%%%%%%%
\begin{abstract}

Financial sentiment analysis is critical for valuation and investment decision-making. Traditional NLP models, however, are limited by their parameter size and the scope of their training datasets, which hampers their generalization capabilities and effectiveness in this field. Recently, Large Language Models (LLMs) pre-trained on extensive corpora have demonstrated superior performance across various NLP tasks due to their commendable zero-shot abilities. Yet, directly applying LLMs to financial sentiment analysis presents challenges: The discrepancy between the pre-training objective of LLMs and predicting the sentiment label can compromise their predictive performance. Furthermore, the succinct nature of financial news, often devoid of sufficient context, can significantly diminish the reliability of LLMs' sentiment analysis. To address these challenges, we introduce a retrieval-augmented LLMs framework for financial sentiment analysis. This framework includes an instruction-tuned LLMs module, which ensures LLMs behave as predictors of sentiment labels, and a retrieval-augmentation module which retrieves additional context from reliable external sources. Benchmarked against traditional models and LLMs like ChatGPT and LLaMA, our approach achieves 15\% to 48\% performance gain in accuracy and F1 score.

%\footnote{\url{}}.

\end{abstract}

%%%%%%%%%%%%%%%%%%%%%%%%%%%%%%%%%%%%%%

\section{Introduction}

Financial sentiment analysis is a critical tool that extracts, quantifies, and studies the affective states and subjective information within financial documents, news articles, and social media content \cite{fsadefinition}. Its significance lies in its potential to forecast market movements and provide valuable insights into investors' behaviors. Given that market reactions are often influenced by news sentiments, which can be positive, negative, or neutral, financial sentiment analysis plays a pivotal role in aiding traders and financial institutions in making informed decisions. It helps manage risks and identify potential investment opportunities by providing a nuanced understanding of the market's emotional undercurrents.

In recent years, numerous studies have turned to Natural Language Processing (NLP) models to enhance the accuracy and efficiency of financial sentiment analysis \cite{araci2019finbert,yang2020finbert,sohangir2018big,day2016deep,wu2023bloomberggpt,yang2023fingpt}. Traditional NLP models, constrained by the limitations of their model parameters and the scale of their training corpora, often lack the capability to comprehensively understand intricate financial news, thereby limiting the efficacy of financial sentiment analysis \cite{araci2019finbert,yang2020finbert,sohangir2018big,day2016deep}. These limitations have sometimes resulted in suboptimal outcomes in financial sentiment analysis tasks. In contrast, the advent of large language models (LLMs) \cite{ouyang2022training,touvron2023llama,wu2023bloomberggpt,yang2023fingpt} has ushered in a new era in the NLP domain. These LLMs, having been pre-trained on vast and diverse corpora, boast formidable zero-shot learning abilities. As a result, they are gradually outperforming many other models across various NLP tasks, owing to their ability to generalize from their extensive training and derive meaningful insights even from previously unseen financial data.

However, directly applying LLMs for financial sentiment analysis poses two notable challenges. Firstly, the discrepancy between the objective function used in LLMs' pre-training and the goal of predicting financial sentiment may result in LLMs' inability to consistently output labels for financial sentiment analysis as expected \cite{thoppilan2022lamda,ouyang2022training}. Secondly, the typical subjects of financial sentiment analysis, such as news flashes and tweets, are characteristically concise and often lack adequate background information. The scarcity of information has not only interfered with the judgment of human experts \cite{malo2014good} but also poses a significant challenge to the accurate prediction of large language models.

To address the aforementioned challenges, in our study, we present a retrieval-augmented large language model framework for financial sentiment analysis. This framework consists of two key components. 1) instruction-finetuned LLMs \cite{ouyang2022training}, which refines LLMs using a limited set of instruction-following examples crafted specifically for financial sentiment analysis, aligning LLMs' predictions with user intentions and significantly boosting their prediction accuracy. 2) retrieval-augmented component \cite{lewis2020retrieval}, which introduces additional context to brief statements from news flashes or tweets. It employs search engines and verified financial sources to gather relevant background information from external sources. This enriched context is then passed to the instruction-tuned LLMs for prediction, resulting in more accurate and nuanced results.

Through extensive evaluations on multiple financial sentiment analysis benchmarks, we demonstrate that compared to traditional smaller-scale sentiment analysis models \cite{araci2019finbert} and general-purpose LLMs, such as ChatGPT \cite{ouyang2022training} and LLaMA \cite{touvron2023llama}, our approach markedly outperforms them.
The primary contributions of this paper can be summarized as follows:

\begin{itemize}
\item We introduce a novel retrieval-augmented large language model framework tailored for financial sentiment analysis. By integrating external knowledge retrieval, we optimize the depth and context of the information feeding into the LLMs, ensuring more nuanced and informed predictions.
\item Our method of instruction tuning leverages a unique set of instruction-following examples. This fine-tuning process realigns LLMs to respond more accurately to user-intended financial sentiment analysis tasks, markedly enhancing their predictive accuracy.
% \item Through extensive evaluations across established benchmarks, we provide empirical evidence that our approach significantly outperforms traditional sentiment analysis models and renowned general-purpose LLMs with 15\% to 48\% performance gain in accuracy and F1 score.
\item Through extensive evaluations on established benchmarks, we demonstrate that our approach outperforms traditional sentiment analysis models and notable general-purpose LLMs, achieving a 15\% to 48\% performance gain in accuracy and F1 score.
\end{itemize}

The remainder of this paper is organized as follows. Section 2 briefly reviews the backgroun and related work. In Section 3, we describe the retrieval augmented method that consists two modules. In Section 4, we present the performance evaluation from three aspects. Section 5 concludes this work and points out directions for future work.

\section{Background and Related Work}

\subsection{Financial Sentiment Analysis}

Financial sentiment analysis has been a prominent area of research in NLP, and deep learning has found widespread application due to its effective feature representation. Early approaches \cite{araci2019finbert,yang2020finbert,sohangir2018big,day2016deep} involved fine-tuning pre-trained models on financial sentiment analysis datasets, but they faced challenges in understanding complex financial news, especially those with numerical information or lacking background context \cite{araci2019finbert}.

Recently, Large Language Models (LLMs) have emerged as an attractive option in NLP. With increasing model size and training data, LLMs demonstrate impressive abilities in in-context learning and chain-of-thought reasoning, allowing them to make predictions in a zero-shot manner. However, LLMs like BloombergGPT \cite{wu2023bloomberggpt} and FinGPT \cite{yang2023fingpt}, tailored for the financial domain, face difficulty in generating expected sentiment labels due to the mismatch between their training objective, typically Causal Language Modeling, and the objective of financial sentiment analysis. Additionally, financial sentiment analysis often deals with brief subjects like news flashes and tweets, lacking sufficient background information. This brevity and contextual deficiency pose a significant challenge for LLMs, making the task of reliable sentiment analysis more difficult.

\subsection{Instruction Tuning}
The latest LLMs such as GPT-3 \cite{brown2020language}, LLaMA \cite{touvron2023llama}, and others have been trained using Causal Language Modeling, which involves predicting the next token given the previous content. However, this training approach introduces randomness in LLMs' outputs, leading to results that may not always align with the desired expectations. 

To address this issue and make LLMs follow specific instructions, researchers have proposed a technique known as instruction tuning \cite{ouyang2022training,lou2023prompt,wei2021finetuned,sanh2021multitask}. It involves fine-tuning pre-trained LLMs on a collection of formatted instances presented in natural language, aiming to guide the LLMs to follow user instructions. These instances typically take the form of task descriptions along with their corresponding desired output, often labeled by humans \cite{ouyang2022training} or semi-automatically constructed \cite{wang2022self}. Through this process, LLMs can be fine-tuned to understand and execute specific instructions effectively, making them more reliable for various applications that require controlled and directed behavior.

\begin{figure*}
\centering
\includegraphics[scale = 0.316]{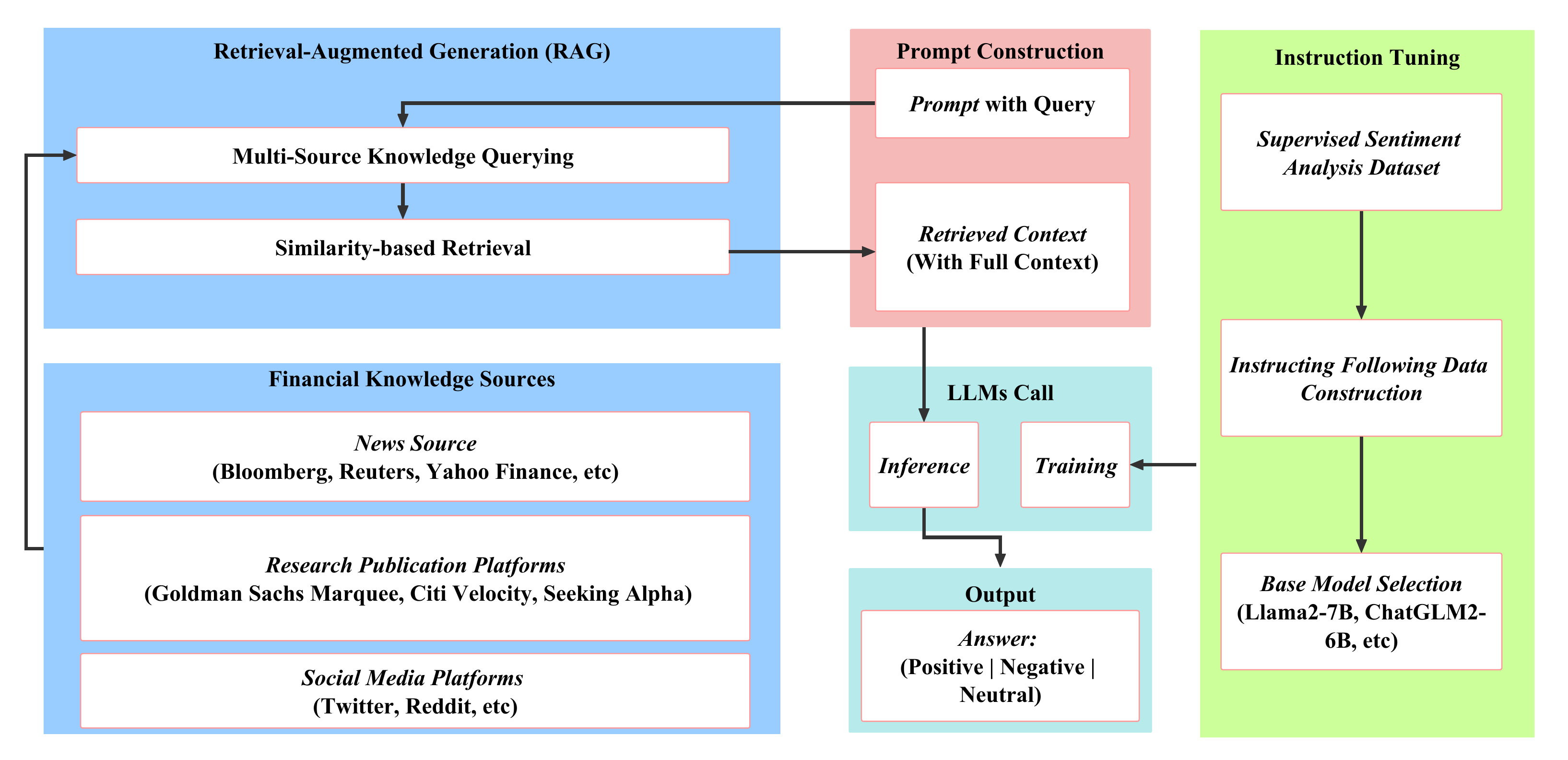}
\caption{Framework of retrieval-augmented large language model for financial sentiment analysis. }
\label{fig:framework2}
\end{figure*}

\subsection{Retrieval Augmented Generation}
% 1. large-scale pretraining context of llms lead to more comprehensive world knowledge. (comparision between the scale of )\\
% 2. numerical awareness. (refers to some reports or articles)\\
% 3. decoder-only vs encoder

Retrieval-augmented generation (RAG) \cite{lewis2020retrieval,cai2022recent} is a technique that combines the strength of context retrieval and LLMs for language generation. RAG operates in a two-step process. First, it retrieves relevant documents using the retrieval module based on the input prompt. These documents are typically sourced from external knowledge bases like news sources, research publications, and social media and provide additional context for the subsequent generation step. Next, the retrieved documents are combined with the original input prompt and fed into the LLMs, generating the final output. The combination of retrieval and generation in RAG allows it to utilize two distinct sources of knowledge: the parametric memory stored in the LLMs' parameters and the nonparametric memory obtained from the corpus of retrieved documents. This dual-knowledge approach enables RAG to effectively guide the generation process and produce more accurate and contextually relevant responses. RAG has been widely used in the areas like open world QA \cite{mao2020generation} and code summarization \cite{liu2020retrieval,parvez2021retrieval}.

% The Retrieval-Augmented Generation (RAG) methodology offers several benefits when applied to tasks involving Large Language Models (LLMs):
% \begin{itemize}
% \item \textbf{Improved Performance on Knowledge-Intensive Tasks}: RAG models have been shown to set the state-of-the-art on several open domain QA tasks, outperforming parametric seq2seq models and task-specific retrieve-and-extract architectures. They generate more specific, diverse, and factual language than a state-of-the-art parametric-only seq2seq baseline \cite{lewis2020retrieval}.
% \item \textbf{Flexibility and Adaptability}: One of the key strengths of RAG is its flexibility. The knowledge it uses can be controlled simply by swapping out the documents it uses for knowledge retrieval. This adaptability is invaluable in situations where facts (or our understanding of the facts) evolve over time \cite{lewis2020retrieval}.
% \item \textbf{Grounding}: RAG is a primary technique for grounding, which is the process of providing LLMs with information that is use-case specific, relevant, and not available as part of the LLM's trained knowledge. Grounding is crucial for ensuring the quality, accuracy, and relevance of the generated output. RAG retrieves information relevant to a task, provides it to the language model along with a prompt, and relies on the model to use this specific information when responding

%\end{itemize}

%%%%%%%%%%%%%%%%%%%%%%%%%%%%%%%%%%%%%%%%%%%%%%%%%%%%%%%%%%

\section{Method}
\subsection{Overview}
Our proposed framework, as shown in Fig. \ref{fig:framework2}, consists two modules of the instruction-tuned LLM and the RAG module. 

In the first module, we apply instruction tuning to fine-tune an open-source pretrained LLM, such as LLaMA \cite{touvron2023llama,touvron2023llama} and ChatGLM \cite{zeng2022glm}, to align their behavior with predicting financial sentiment labels when provided with financial news or tweets. This process involves constructing an instruction-following dataset specific to the task of financial sentiment analysis and using it to fine-tune the pretrained LLM.

The RAG module plays a crucial role in the framework by retrieving pertinent background information from external sources related to the input query. These external sources include well-authenticated news platforms like Bloomberg and Reuters, research publications from institutions like Goldman Sachs and Citi Velocity, and social media platforms such as Twitter and Reddit. We employ a multi-source query and similarity-based retrieval approach to locate the most relevant information from these sources.

Subsequently, the retrieved context is combined with the original query to construct the final query. The instruction-tuned LLM is then called upon to generate a sentiment prediction based on this augmented query. In this way, the missing background knowledge is provided to the LLM, enabling it to make more accurate predictions.

The implementation of these steps will be further detailed in the following sections, showcasing how our framework effectively incorporates instruction tuning and retrieval-augmented generation to enhance the accuracy and precision of financial sentiment analysis.

\subsection{Instruction-tuned LLMs\label{sec:in_tuning}}

% We adopt the instruction tuning method of an LLM on financial sentiment analysis datasets. 

% Instruction tuning is an efficient method of adapting pre-trained LLMs to downstream tasks without having to train the models from scratch. This greatly reduces the time and computational resources required to train models. Based on the instruction tuning pipeline, we can verify that even based on open-source models, we can achieve or exceed the performance of closed-source models at a small cost. This means that we do not need to invest a lot of resources to develop and train our models. Instead, we can use existing open-source models and tools, adapt them through instruction tuning, and achieve excellent results. This process is divided into three main steps:

Instruction tuning proves to be a highly effective approach to align the behavior of LLMs with user instructions, particularly in our study, where we aim to predict financial sentiment labels. Encouraging results from recent studies \cite{alpaca,vicuna2023,zhang2023instruct} demonstrate that limited instruction following data, when used in instruction tuning, allows the resulting LLMs to adhere remarkably well to user instructions.

There are typically three steps to apply instruction tuning in the domain of financial sentiment analysis. Firstly, we construct an instruction-following dataset, consisting of paired instructions and their corresponding expected responses – essentially the labels of sentiment. This dataset serves as the foundation for guiding the LLMs to understand the user's instructions effectively. The second step involves fine-tuning the LLMs on the constructed dataset. Through this fine-tuning process, the model learns to generate the expected response accurately when provided with instructions to predict sentiment labels. The final step is to map the generated outputs from the LLMs back into predefined sentiment classes. This step further aligns the prediction with predefined sentiment classes and allows the model's performance to be measurable. We detail these steps in the following.

\subsubsection{Formatting Financial Sentiment Instruction Following Dataset}

\begin{figure*}[ht]
\centering
\includegraphics[scale = 0.34]{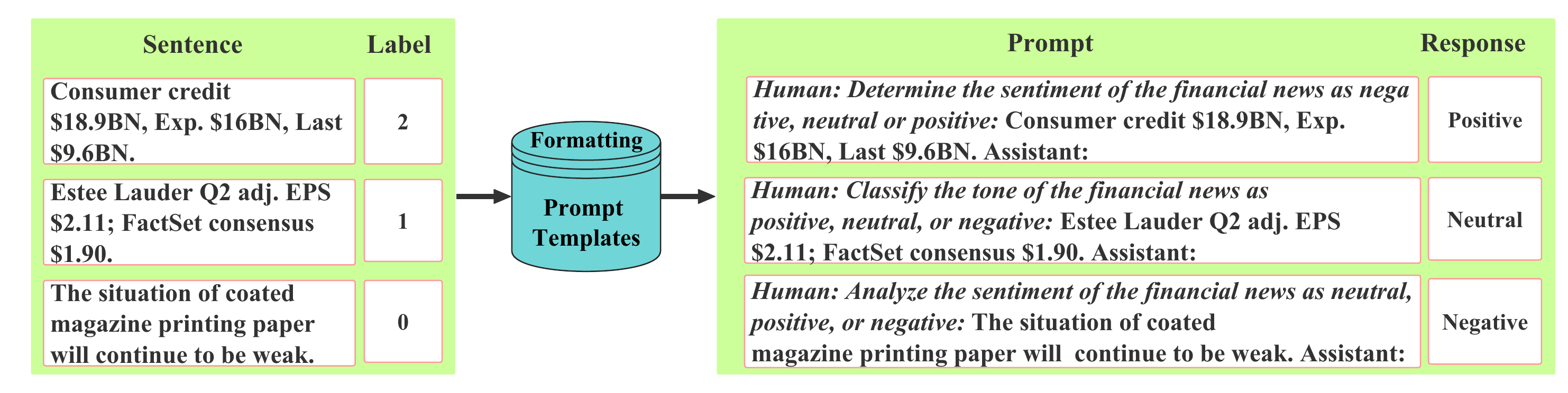}
\caption{Formatting sentiment analysis dataset into instruction-following dataset.}
\label{fig:framework}
\end{figure*}

Creating a financial sentiment instruction-following dataset through manual labeling requires the expertise of specialized financial professionals, which can be costly. An alternative approach is to convert existing supervised financial sentiment analysis datasets into instruction-following datasets at a lower cost \cite{wang2022self,wei2021finetuned,ouyang2022training,zhao2023survey}. These datasets are often formatted as text classification tasks where the \textbf{inputs} are the financial news or headlines and the \textbf{outputs} are integer-type labels representing \textit{positive}, \textit{negative} and \textit{neutral} sentiments. 

Following \cite{zhao2023survey}, we create 10 human-written \textbf{instructions} describing the task of financial sentiment analysis, and formulate each sample from the original dataset by combining one randomly selected \textbf{instruction} with the \textbf{input} and \textbf{output} in the format of "Human: [\textbf{instruction}] + [\textbf{input}], Assistant: [\textbf{output}]". This process is shown in Fig. \ref{fig:framework}. %In this way,  the model is instructed to generate whether the sentiment is positive, neutral, or negative.
%prompt 10个每次randomly选一个防止overfit

\subsubsection{Instruction Tuning}

Instruction tuning involves fine-tuning pre-trained LLMs using the instruction-following dataset. We tokenize the texts into tokens with the byte-pair encoding (BPE) algorithm \cite{sennrich2015neural} first. Then we fine-tune the LLMs with the causal language modeling (CLM) objective, which maximizes the likelihood of predicting the next token in a sequence given the preceding context. This process is achieved by minimizing the following negative log-likelihood,
\begin{equation}
    % L(\theta) = - \sum \log P(x_i | x_1, x_2, \ldots, x_{i-1}; \theta)
    \mathcal{L}_{\text{CausalLM}} = - \sum_{t=1}^{T} \log P(w_t | w_1, w_2, \ldots, w_{t-1}; \theta),
\end{equation}
where $T$ is the length of the input sequence, $w_i$ represents the $i$-th token in the sequence, $\theta$ denotes the model parameters, and $P(w_i | w_1, w_2, \ldots, w_{i-1}; \theta)$ represents the conditional probability of predicting the target token $x_i$ given the preceding tokens $x_1, \ldots, x_{i-1}$. By optimizing this objective function, the model learns to maximize the probability of generating the expected financial sentiment label given the instruction. We use the gradient-based optimization method to minimize the above objective where the gradient is represented as,
\begin{equation}
    \nabla_{\theta} \mathcal{L}_{\text{CausalLM}} = - \sum_{t=1}^{T} \frac{\partial \log P(w_t | w_1, w_2, \ldots, w_{t-1}; \theta)}{\partial \theta}.
\end{equation}
This gradient is typically computed by backpropagation through time \cite{werbos1988generalization}. It involves computing the gradients at each time step 
$t$ and then propagating them back through the entire sequence of time steps.

\subsubsection{Mapping the Generated Outputs into Sentiment Classes}
% Post fine-tuning, we map the model-generated text into sentiment labels. The generation task output (positive, neutral, negative) is interpreted as sentiment labels. The capacity of the LLM is thus effectively channeled towards the accurate determination of financial sentiment.

% Our instruction tuning approach applied to LLMs addresses the limitations observed in existing models, including FinBERT. It improves the accuracy of financial sentiment analysis, better leveraging the contextual understanding and numerical sensitivity of the model.

Since the instruction-finetuned LLM is an autoregressive generative model, even though we train it using instruction-following dataset to guide its output towards the desired sentiment label, it still has the possibility of generating free-style text. Therefore, we need to map the model's output back to the specified three emotions for proper evaluation. Our approach is as follows: 
% if the model's output contains "positive," "negative," or "neutral" terms, we map it to the corresponding label; 
we sequentially check if the output results contain "negative", "neutral", or "positive". Once a term is found, we map it to the corresponding label.
Otherwise, we consider it a "neutral" sentiment.

\subsection{RAG Module}
% In this work, our primary objective is to explore the adaptation of domain-specific retrieval augmentation for financial applications.

% The RAG procedure we developed consists of three main components: (a) specifications of financial knowledge sources, (b) an external knowledge base, and (c) its retrieval system.
RAG is an effective method for injecting external knowledge into LLM to enhance the accuracy of response generation. The implementation of the RAG module involves several steps. Firstly, we set up external knowledge sources which are highly likely to contain relevant financial background information. Next, we perform a two-step knowledge retrieval process, consisting of Multi-Source Knowledge Querying and Similarity-Based Retrieval. These steps enable us to gather relevant context related to the input query. Finally, we combine the original input query with the retrieved context, creating input data for the instruction-tuned LLM, which generates the final result.

\subsubsection{Setup External Knowledge Sources}
% We propose a method to progressively build an external knowledge base suitable for financial decision-making. This involves not only gathering information from various sources but also organizing and analyzing this information to create a comprehensive and actionable knowledge base.
% For any information retrieval pertinent to financial decision-making, a source must satisfy a few principles: (1) Consistency, (2) authentication, (3) relevance, (4) completeness, and (5) insight \cite{grootArticle,grootBook}. Consistency ensures no contradictions within the material and alignment with other sources. Authentication verifies the credibility of the information source. Relevance ensures the information is timely and useful. Completeness requires the information to provide a comprehensive view of a financial event. Insight demands the information to provide meaningful analysis for practical applications. We consider the following three kinds of data sources of external knowledge:

When retrieving relevant financial context based on the query, our objective is to access authentic, pertinent, insightful, and comprehensive data, as opposed to random internet searches. To achieve this, we first identify the following sources of information:

\paragraph{News Sources} Esteemed outlets such as Bloomberg, Yahoo Finance, Reuters, CNBC, and Market Screener supply information that is inherently consistent and crucial for financial interpretations. These sources tend to have stringent internal guidelines for their writers and reporters, ensuring reliable and verified content. Furthermore, owing to the nature of their operations, these outlets often offer the earliest reports on a variety of financial news.

\paragraph{Research Publication Platforms} Centralized, as well as crowd-based research publication platforms, provide a wealth of financial insights.
\begin{itemize}
\item\textbf{Centralized Publishers} Renowned institutions such as Goldman Sachs and Citi proffer exclusive research services, Marquee and Velocity respectively, predominantly to their institutional clients. Given their direct applicability, these researches provide highly consistent, systematic, and authenticated insights.
\item\textbf{Crowd-based Publishers} Platforms like Seeking Alpha serve as a repository for diverse insights from independent contributors. They cover a broad spectrum of financial information that includes a vast array of price movement analyses, transcripts of earnings calls and conferences, and investment research pertaining to companies of all sizes.
\end{itemize}

All of these sources provide retrieval APIs, enabling us to access and retrieve information.

% \begin{itemize}
% \item\textbf{Centralized Publishers}: These include equity research publications from sell-side financial institutions like Goldman Sachs Marquee and Citi Velocity. These sources are often exclusive to key clients and provide internally consistent, well-authenticated, and insightful information. However, their global consistency with crowd-based publishers is often scrutinized and adjusted by analysts to ensure a more objective market view. While these sources provide in-depth analysis and forecasts, their predictions may be influenced by their institutional interests.

% \item\textbf{Crowd-Based Publishers}:  Platforms such as Seeking Alpha are known for their insightful content, thanks to a marketplace of named individual writers. These platforms offer a diversity of opinions and analyses, providing a comprehensive overview of market insights. However, the consistency and authentication of these sources can vary significantly due to the lack of stringent editorial standards like those in traditional news outlets or research publications.
% \end{itemize}

\paragraph{Social Media Platforms:} Social media platforms like Twitter and Reddit have become significant sources of financial information. These platforms offer real-time updates and discussions, which can be insightful for understanding market sentiment and trends. However, the information on these platforms can be highly volatile and unverified, necessitating careful analysis and cross-referencing with other sources.

\subsubsection{Two-Step Knowledge Retrieval}

% \paragraph{Trigger} The process commences with a trigger, which could be a user query or instruction.

We retrieve contextual financial information for a given query through a two-step process.

\paragraph{Multi-Source Knowledge Query} 
% The financial news, pre-processed into text and optional parameters of urls and tickers, is queried on a search engine with specified query parameters. 

Financial news headlines or tweets are typically short and often include irrelevant content like tickers. To address this, our first step involves using regular expressions to preprocess the text and remove irrelevant tickers or symbols. Subsequently, we utilize various knowledge sources' retrieval APIs to extract relevant information. If the news item contains time information, we perform searches within that specific time range. The search returns a list of relevant context snippets from identified financial sources. For each context snippet, we gather the original headline, editorial bullet points, article body paragraphs, posts, and reposts as the full context. This query strategy allows us to capture a wide spectrum of information related to financial news.

\paragraph{Similarity-Based Retrieval}
Even after the initial retrieval, the content obtained may still contain a considerable amount of irrelevant information, which could potentially hinder sentiment prediction accuracy. To address this issue, we propose an advanced retrieval algorithm based on similarity. This algorithm aims to further filter and extract the most relevant content from the results obtained in the first step. Specifically, 
we use a modified overlap coefficient as the similarity measure for retrieval and empirically select those context with a similarity higher than 0.8 to the input query. The overlap coefficient, also known as the Szymkiewicz-Simpson coefficient \cite{similarityScore}, is used to measure the similarity degree between two samples. In the task of sentence-context pair similarity evaluation, this coefficient measures the number of words in the intersection divided by the union of the pair. The specific formula given as follows:
\begin{equation}
    \text{overlap}(\mathbf{X}, \mathbf{Y}) = \frac{|\mathbf{X} \cap \mathbf{Y}|}{min(|\mathbf{X}|, |\mathbf{Y}|)},
    \label{eql:overlap}
\end{equation}
where ${\mathbf{X}}$ and ${\mathbf{Y}}$ represent sets of financially relevant tokens from a queried sentence and its context, respectively.

% We prefer this coefficient over other term-based (such as Jaccard) and similarity scores based on characters, knowledge, and corpus for two major reasons: (1) Financial news often contains links and tickers that whose absence and differences among the contrasting sentences should not indicate greater discrepancy and (2) financial contexts are often longer than the queried statement and have varied lengths themselves, to the extent that a union set’s length anywhere in the formula may counterweigh a large value of the intersection. The overall two-step knowledge retrieval algorithm is shown in Algo. \ref{alg:generator}.

We prefer the Szymkiewicz-Simpson coefficient over the semantic similarity for two major reasons. In financial news, the need for exact matches, especially for tickers, is paramount. This coefficient emphasizes this hard matching, minimizing irrelevant retrievals. In contrast, semantic similarity can sometimes miss the intricacies of specific financial terms. Moreover, the challenge of short-to-long text matching, highlighted in \cite{rawte2020comparative}, is adeptly managed by the Szymkiewicz-Simpson coefficient, ensuring relevant news are not overshadowed by length. The overall two-step knowledge retrieval algorithm is shown in Alg. \ref{alg:generator}.

% We prefer this coefficient over the semantic similarity for two major reasons: (1) Financial news often contain links and tickers that whose absence and differences among the contrasting sentences should not indicate greater discrepancy and (2) financial contexts are often longer than the queried statement and have varied lengths themselves, to the extent that a union set’s length anywhere in the formula may counterweigh a large value of the intersection. The overall two-step knowledge retrieval algorithm is shown in Algo. \ref{alg:generator}.

% This process ensures that the most pertinent and reliable information is incorporated into our knowledge base. 

% \begin{algorithm}[H]
% \caption{Financial Information Multi-Source Retrieval System}
% \label{alg:generator}

% \function{\textit{generate}}{Object pivot}
% \REQUIRE Query $\mathbf{Q}$
% \ENSURE Context $\mathbf{C}$
% \FOR{each query $\mathbf{Q}$}
%     \STATE Obtain posting list, p, of documents, each of which containing some tokens q in $\mathbf{Q}$
%     \FOR{each document $\mathbf{D}$ in p}
%         \IF{${\text{overlap}(\mathbf{Q},\mathbf{D})} > 0.8$}
%             \STATE Scan document $\mathbf{D}$ and parse into semantic groups separated by separative tokens (paragraphs, bullet points)
%             \FOR{each semantic group $\mathbf{G}$}
%                 \IF{${\text{overlap}(\mathbf{Q},\mathbf{G})} > 0.7$}
%                     \STATE Concatenate $\mathbf{G}$ to context $\mathbf{C}$
%                 \ENDIF
%             \ENDFOR
%         \ENDIF
%     \ENDFOR
% \ENDFOR
% \end{algorithm}

\begin{algorithm}[h]
\caption{Financial Knowledge Retrieval}
\label{alg:generator}
\begin{algorithmic}[1]
\REQUIRE Query $\mathbf{Q}$
\ENSURE Context $\mathbf{C}$
    \STATE Search $\mathbf{Q}$ using a search engine and obtain a list of documents, $\mathbf{D}$, where each contains some phrases of $\mathbf{Q}$
    \STATE $\mathbf{C} \leftarrow \emptyset$
    \FOR{document $\mathbf{d} \in \mathbf{D}$}
        \IF{${\text{overlap}(\mathbf{Q},\mathbf{d})} > 0.8$}
            \STATE Split $\mathbf{d}$ into syntactic units, $\mathbf{u}_i$ for ${\{i | 1 \leq i \le n + 1\}}$, which are separated by $n$ separative tokens (i.e., paragraphs, bullet points)
            
            \FOR{ $i = 1$ to $n+1$}
                \IF{${\text{overlap}(\mathbf{Q},\mathbf{u}_i)} > 0.7$}
                    \STATE $\mathbf{C}$=Concat($\mathbf{C}$, $\mathbf{u}_i$)
                \ENDIF
            \ENDFOR
        \ENDIF
    \ENDFOR
\end{algorithmic}
\end{algorithm}
% \vspace{-3mm}

% \subsubsection{Overlap Similarity Score}

%\paragraph{Summarization} Given the accuracy and relevance principles of financial knowledge, we must first summarize the contexts. We use a model such as GPT-3.5 Text-Davinci-003 to summarize contexts with the prompt: “Given the contextual snippets below, current context: [contextual snippets]; past contexts that may be relevant: [contextual snippets]. Summarize their contents and their sentiments.” We then use the summarized context as part of the information context for the financial statement. This process allows us to distill the most important information from the knowledge base and present it in a concise and understandable format.

\section{Performance Evaluation}

In this section, we evaluate the effectiveness of instruction fine-tuning and RAG. To validate our method's performance, we compare it against state-of-the-art sentiment analysis models and the general-purpose LLMs. Our experimental results validate the effectiveness of our approach. With only a small amount of instruction-following data, our model consistently outperforms other baselines in sentiment analysis and its performance can be further enhanced with the RAG module. The code of this experiment is available at Github\footnote{\url{https://github.com/AI4Finance-Foundation/FinGPT/tree/master/fingpt/FinGPT_RAG/instruct-FinGPT}}.

\subsection{Datasets}
%分一下数据集

\subsubsection{Training Datasets}
Our training data is an amalgamation of the Twitter Financial News dataset \cite{twitter2022finance} and FiQA dataset \cite{fiqa}, resulting in a collection of $10,501$ samples.
\begin{itemize}
    \item \textbf{Twitter financial news sentiment training}: This dataset \footnote{\url{https://huggingface.co/datasets/zeroshot/twitter-financial-news-sentiment}} is a corpus of news tweets that pertain to the financial sector. Its primary purpose is the classification of financial sentiment within the context of Twitter discussions. The dataset comprises 9,540 samples for training, each annotated with one of three labels: Bearish, Bullish, or Neutral. 
    \item \textbf{FiQA dataset}: This dataset \footnote{\url{https://huggingface.co/datasets/pauri32/fiqa-2018}} includes 961 samples. Each sample has been annotated with one of three labels: positive, neutral, or negative, denoting the sentiment conveyed in the corresponding text. 
\end{itemize}

\subsubsection{Testing Datasets}

\begin{itemize}
    \item \textbf{Twitter financial news sentiment validation (Twitter Val)}: This is the validation split of the Twitter dataset which contains 2,388 samples. It can validate how well the model can predict the financial sentiment from the social media. It's important to note that the platform often lacks clear sources and context for news items.
    \item \textbf{Financial PhraseBank (FPB) dataset}: This dataset \footnote{\url{https://huggingface.co/datasets/financial_phrasebank}} \cite{malo2014good} comprises 4,840 samples randomly extracted from financial news articles available on the LexisNexis database. The samples were carefully annotated by a team of 16 annotators with backgrounds in finance and business, ensuring high quality annotations.
\end{itemize}

For all the above datasets, we use the approach mentioned in Section \ref{sec:in_tuning} to format them as instruction-following datasets before training and testing.

%The choice to use Twitter for financial news sentiment validation, or "Twitter Val," is due to the platform's real-time and diverse nature. Despite the lack of clear sources for news items on Twitter, our aim is to trace these back to their origins, often leading to reputable outlets like Seeking Alpha, Bloomberg, and Reuters. This allows for reliable sentiment validation and informed financial decisions.

\subsection{Model Training}

% The training parameters are given in Table \ref{tab:parameters}. 
We initialize our model with Llama-7B and perform instruction tuning over $10$ epochs. The training process utilizes the AdamW optimizer \cite{loshchilov2017fixing}, with a batch size of $32$, an initial learning rate of $1e^{-5}$, and a weight decay of $0.1$. To maintain efficiency, we set a maximum input text length of $512$ tokens. We utilize DeepSpeed \cite{deepspeed} for the fine-tuning process on 8$\times$A100 (40GB) GPUs, resulting in a total training time of $58$ minutes.

% \begin{table}[t]
% \centering
% \begin{tabular}{ll}
% \hline
% Parameters        & Value           \\ \hline
% Learning rate    & 1e-5            \\
% Weight decay     & 0.1           \\
% Batch size       & 32              \\
% Training epochs  & 10              \\
% LR scheduler     & CosineAnnealing \\
% Num warmup steps & 0               \\
% Max token length & 512             \\ 
% GPUs             & 8 $\times$ A100 (40GB)               \\

% \hline
% \end{tabular}
% \caption{Hyper-parameters for training.}
% \label{tab:parameters}
% \vspace{-5mm}
% \end{table}

\subsection{Baseline Models}

\paragraph{BloombergGPT} \cite{wu2023bloomberggpt} BloombergGPT is a 50 billion parameter language model that is trained on a wide range of financial data. As it's a closed-source model, we directly use their reported performance on the FPB dataset.

\paragraph{ChatGPT} \cite{ouyang2022training} ChatGPT is the cutting-edge closed-source LLM developed by OpenAI. The use of OpenAI's ChatGPT API for sentiment analysis involves four steps: API setup, data preparation using the instruction-following dataset, making requests using the GPT-4.0 API, and interpreting the direct sentiment output from the response.

% \paragraph{Llama-7B} \cite{touvron2023llama} Llama-7B is an open-source LLM developed by Meta where most of the training corpus is in English. We obtained the Llama-7B\footnote{We use Llama-7B for research and education purposes.} model from Meta and used it for zero-shot inference, keeping the same inference setting as our base model.

\paragraph{Llama-7B } \cite{touvron2023llama} Llama-7B is an open-source LLM created by Meta, with the majority of the training corpus being in English. We acquired the Llama-7B\footnote{We utilize Llama-7B for research and education purposes.} model from Meta. %  and maintain the identical inference configuration as our base model

% \paragraph{ChatGLM2-6B} \cite{zeng2022glm} ChatGLM2-6B is an open-source LLM developed by Tsinghua University which supports both English and Chinese. We obtained the ChatGLM2-6B model from Hugging Face Model Hub and used it for zero-shot inference.

\paragraph{ChatGLM2-6B} \cite{zeng2022glm} ChatGLM2-6B is an open-source LLM crafted by Tsinghua University, supporting both English and Chinese. We acquired the ChatGLM2-6B model from Hugging Face Model Hub.

\paragraph{FinBERT} \cite{araci2019finbert} FinBERT is a financial sentiment analysis model which is fine-tuned on the pretrained BERT language model. The FinBERT model is also accessible through the Hugging Face Model Hub.

\subsection{Evaluation and Analysis}

To evaluate the performance of our model, we first compare our instruction-tuned LLM with the sentiment analysis model FinBERT and the general-purpose LLM to verify the effectiveness of the instruction tuning. We then compare the LLMs including ours and the baselines with and without RAG to further validate the efficacy of RAG.

\subsubsection{Performance metrics}

The performance metrics for our model include accuracy, and F1-score. Accuracy measures the proportion of correct predictions, 
% precision assesses the proportion of true positive predictions, recall (or sensitivity) measures the ability of the model to find all relevant instances, 
and the F1-score represents the harmonic mean of precision and recall.

\subsubsection{Assessment of Instruction Finetuning}

% We evaluate the zero-shot ability of our model, which refers to how well the model can generalize to other unseen financial datasets. A model with strong zero-shot capabilities can provide more robust and versatile results in real-world applications. We compare our model with BloombergGPT, ChatGPT4.0, ChatGLM2-6B, and Llama-7B on the  FPB dataset with the same testing set as BloombergGPT. We excluded FinBERT from this comparison as it uses FPB as its training dataset.

In this experiment, we aim to verify the effectiveness of instruction tuning, denoted as "Ours" in the presented Table \ref{table:zero-shot}. The comparative analysis is performed against all the baseline models. 
% with FinBERT being recognized as the state-of-the-art (SOTA) model for financial supervised sentiment analysis, while the other models, such as BloombergGPT, ChatGLM2-6B, Llama-7B, and ChatGPT 4.0, are acknowledged as popular general-purpose LLMs. 
The evaluation is conducted on the Financial PhaseBank (FPB) and Twitter Val. We excluded FinBERT from the comparison on FPB as it uses the exact dataset for training.

\begin{table}[ht]
\centering
\begin{tabular}{lcccc}
\hline
Dataset & \multicolumn{2}{c}{FPB} & \multicolumn{2}{c}{Twitter Val} \\ \hline
Metrics         & Acc   & F1     & Acc   & F1     \\ \hline
FinBERT \cite{araci2019finbert}  & - & - & 0.725  & 0.668  \\
BloombergGPT \cite{wu2023bloomberggpt}    & -        & 0.510    & -        & -        \\
ChatGLM2-6B \cite{zeng2022glm}    & 0.474     & 0.402   & 0.482     & 0.381     \\
Llama-7B \cite{touvron2023llama}      & 0.601     & 0.397    & 0.544     & 0.363     \\
ChatGPT 4.0 \cite{ouyang2022training}     & 0.643     & 0.511    & 0.788 & 0.652       \\
Ours            & \textbf{0.758} & \textbf{0.739}   & \textbf{0.863} & \textbf{0.811}           \\
\hline
\end{tabular}
\caption{Comparison between our model and the baselines on the datasets of financial phaseBank (FPB) and Twitter Val.}
\label{table:zero-shot}
\end{table}

\begin{table}[bt]
\centering
\begin{tabular}{lll}
\hline
Metrics          & Acc   & F1          \\ \hline
% FinBert \cite{araci2019finbert}    & 0.725  & 0.668            \\
% ChatGLM2-6B  \cite{zeng2022glm}    & 0.482     & 0.381         \\
% Llama-7B \cite{touvron2023llama}    & 0.544     & 0.363         \\
ChatGPT 4.0 w/o RAG & 0.788 & 0.652    \\
ChatGPT 4.0 w/ RAG & \textbf{0.813} &\textbf{0.708} \\ \hline
Ours w/o RAG      & 0.863 & 0.811              \\
Ours w/ RAG  & \textbf{0.881} & \textbf{0.842}        \\
\hline
\end{tabular}
\caption{Experimental results on the Twitter Val dataset.}
\label{tab:booktabs}
\end{table}

% \begin{table}[ht]
% \centering
% \begin{tabular}{lll}
% \hline
% Metrics         & Accuracy   & F1          \\ \hline
% BloombergGPT \cite{wu2023bloomberggpt}    & -        & 0.51            \\
% ChatGPT 4.0 \cite{ouyang2022training}     & 0.64     & 0.51           \\
% ChatGLM2-6B \cite{zeng2022glm}    & 0.47     & 0.40         \\
% Llama-7B \cite{touvron2023llama}      & 0.60     & 0.40         \\
% Ours            & \textbf{0.76} & \textbf{0.74}              \\
% \hline
% \end{tabular}
% \caption{Zero-shot evaluation between BloombergGPT, general-purpose LLMs like ChatGPT, and our model on the dataset of financial phaseBank (FPB).}
% \label{table:zero-shot}
% \vspace{-7mm}
% \end{table}

% The outcomes in Table \ref{table:zero-shot} of the experiment highlight the superior performance of our instruction tuning approach over the baseline models. Our model achieved the highest scores across both datasets and on both metrics, marking a significant improvement in performance. 

The outcomes in Table \ref{table:zero-shot} suggest that our instruction-tuned Llama-7B model outperforms the others, achieving the highest accuracy and F1 score. The process of fine-tuning with instruction-following data enhances the model's ability to discern sentiment in financial phrases, leading to superior performance compared to both ChatGPT 4.0 and the original Llama-7B model. From these findings, it is evident that the instruction tuning method significantly improves the model's performance on financial sentiment analysis.

\begin{table}[hbt]
\centering
\begin{tabular}{lp{4.85cm}l}
\hline
          & Text   & Result          \\ \hline
w/o RAG & \$ENR - Energizer shakes off JPMorgan’s bear call. & Neutral    \\ \hline
w/ RAG & \textit{"Energizer shakes off JPMorgan's bear call. JPMorgan \textbf{hikes Energizer Holdings (NYSE:ENR) to a Neutral rating from Underweight}... We came away \textbf{encouraged} by some of the company's initiatives and believe their focus on innovation and brand investment can lead to relative outperformance going forward... Shares of Energizer are \textbf{0.46\% premarket to \$50.44}."} & \textbf{Positive} \\ \hline
\end{tabular}
\caption{Case study: before and after using RAG.}
\label{tab:case}
\end{table}

\subsubsection{Performance of RAG Module}
We verify the effectiveness of RAG module on both our instruction-tuned model and the ChatGPT 4.0 on the Twitter Val dataset in this experiment. From the results presented in Table \ref{tab:booktabs},
it demonstrates the introduction of RAG context will universally improve the performance of the LLMs which verifies that the retrieved context enhances the information and allow the LLMs to make more accurate prediction. Specially, our model with RAG again achieves the best performance among all the methods.

% However, it is worth noting that both Llama-7B and our model exhibit slower inference speeds compared to FinBERT. This is due to the significantly larger model parameters and FLOPs of Llama-7B and our model compared to FinBERT. The increased model size and complexity of these models necessitate more computational resources and processing time. Despite this, Llama-7B and our model, due to their shared model architecture, exhibit identical testing speeds.

% \paragraph{Example of using RAG to retrieve context}
To better highlight the RAG module's effectiveness, we present a case study in Table \ref{tab:case}. Initially, the statement's ambiguity causes our instruction-tuned model to misclassify it as "neutral." With RAG, we augment the context using information from Seeking Alpha, clarifying the phrase "shakes off" to indicate a rating upgrade, which helps our model correctly reclassify the statement as "positive." This showcases RAG's ability to enhance model comprehension and provide a more nuanced understanding of the sentiment in the news headline.

\section{Conclusion and Future Work}

% In conclusion, this paper presents a novel retrieval augmented large language model framework specifically tailored for financial sentiment analysis. Our unique method of instruction tuning has realigned LLMs to respond more accurately to user-intended financial sentiment analysis tasks, significantly enhancing their predictive accuracy. By integrating external knowledge retrieval, we have been able to enhance the depth and context of the information feeding into the LLMs, ensuring more nuanced and informed predictions. 

In conclusion, this paper unveils a novel retrieval-augmented large language model framework tailored for financial sentiment analysis. Our unique instruction tuning method has realigned LLMs to respond more accurately to user-intended financial sentiment analysis tasks, significantly enhancing their predictive accuracy. The integration of external knowledge retrieval has further enriched the depth and context of the information fed into the LLMs, enabling more nuanced predictions.

However, a limitation of our approach is its exclusive reliance on textual similarity to retrieve relevant information. This method overlooks crucial macroeconomic information related to the timing of the news and microeconomic information concerning the financial and operational status of the related enterprise. Incorporating such economic data could provide a more holistic view, allowing LLMs to make more accurate judgments. 
Future work could explore amalgamating these additional economic dimensions with textual data, to further improve the precision and reliability of financial sentiment analysis performed by large language models.

% Future work could incorporate our framework into real markets such as the components of the S\&P 500, to further showcase the versatility and effectiveness of our approach.

%\textbf{Disclaimer: We are sharing codes for academic purposes under the MIT education license. Nothing herein is financial advice, and NOT a recommendation to trade real money. Please use common sense and always first consult a professional before trading or investing.}

\bibliographystyle{IEEEbib}
\bibliography{references.bib}

\end{document}